\pdfoutput=1
\documentclass[10pt,twocolumn,letterpaper]{article}
 
\usepackage{cvpr}
\usepackage{times}
\usepackage{epsfig}
\usepackage{graphicx}
\usepackage{amsmath}
\usepackage{amssymb}
\usepackage{booktabs}
\usepackage{mathtools}
\usepackage{subcaption}
\usepackage{color}
\usepackage{commath}
\usepackage{verbatim}

\usepackage[pagebackref=true,breaklinks=true,letterpaper=true,colorlinks,bookmarks=false]{hyperref}

\newcommand{\mX}{\ensuremath{X}}
\newcommand{\mx}{\ensuremath{x}}
\newcommand{\mhatX}{\ensuremath{\hat \mX}}
\newcommand{\mL}{\ensuremath{\mathcal L}}
\newcommand{\my}{\ensuremath{y}}
\newcommand{\mhaty}{\ensuremath{\hat y}}
\newcommand{\mxi}{\ensuremath{\xi}}
\newcommand{\mXi}{\ensuremath{\Xi}}
\newcommand{\mQ}{\ensuremath{Q}}
\newcommand{\msetTheta}{\ensuremath{\Theta}}
\newcommand{\mtheta}{\ensuremath{\theta}}

\newcommand{\mphi}{\ensuremath{\phi}}
\newcommand{\mIm}{\ensuremath{I}}

\newcommand{\msimp}{\ensuremath{\Delta}}
\newcommand{\mB}{\ensuremath{B}}
\newcommand{\mP}{\ensuremath{P}}
\newcommand{\mtau}{\ensuremath{\tau}}
\newcommand{\ourMethod}{RAWKS} 

\DeclarePairedDelimiterX{\kldx}[2]{(}{)}{%
  #1\;\delimsize\|\;#2%
}
\newcommand{\kld}{\textrm{KL}\kldx}

\cvprfinalcopy 

\def\cvprPaperID{3380} 

\ifcvprfinal\fi
\begin{document}

\title{Learning random-walk label propagation \\for weakly-supervised semantic segmentation}

\author{Paul Vernaza \qquad Manmohan Chandraker\\
NEC Laboratories America, Media Analytics Department\\
10080 N Wolfe Road, Cupertino, CA 95014\\
{\tt\small \{pvernaza,manu\}@nec-labs.com}
} 

\maketitle

\begin{abstract}
  Large-scale training for semantic segmentation is challenging due to
  the expense of obtaining training data for this task relative to
  other vision tasks.  We propose a novel training approach to address
  this difficulty.  Given cheaply-obtained sparse image labelings, we
  propagate the sparse labels to produce guessed dense labelings.  A
  standard CNN-based segmentation network is trained to mimic these
  labelings.  The label-propagation process is defined via random-walk
  hitting probabilities, which leads to a differentiable
  parameterization with uncertainty estimates that are incorporated
  into our loss.  We show that by learning the label-propagator
  jointly with the segmentation predictor, we are able to effectively
  learn semantic edges given no direct edge supervision.  Experiments
  also show that training a segmentation network in this way
  outperforms the naive approach. \footnote{This article is a corrected version of an article published in CVPR 2017: \url{https://doi.org/10.1109/CVPR.2017.315}}
\end{abstract}

\section{Introduction}

\begin{figure}
  \centering
  \includegraphics[width=2.75in]{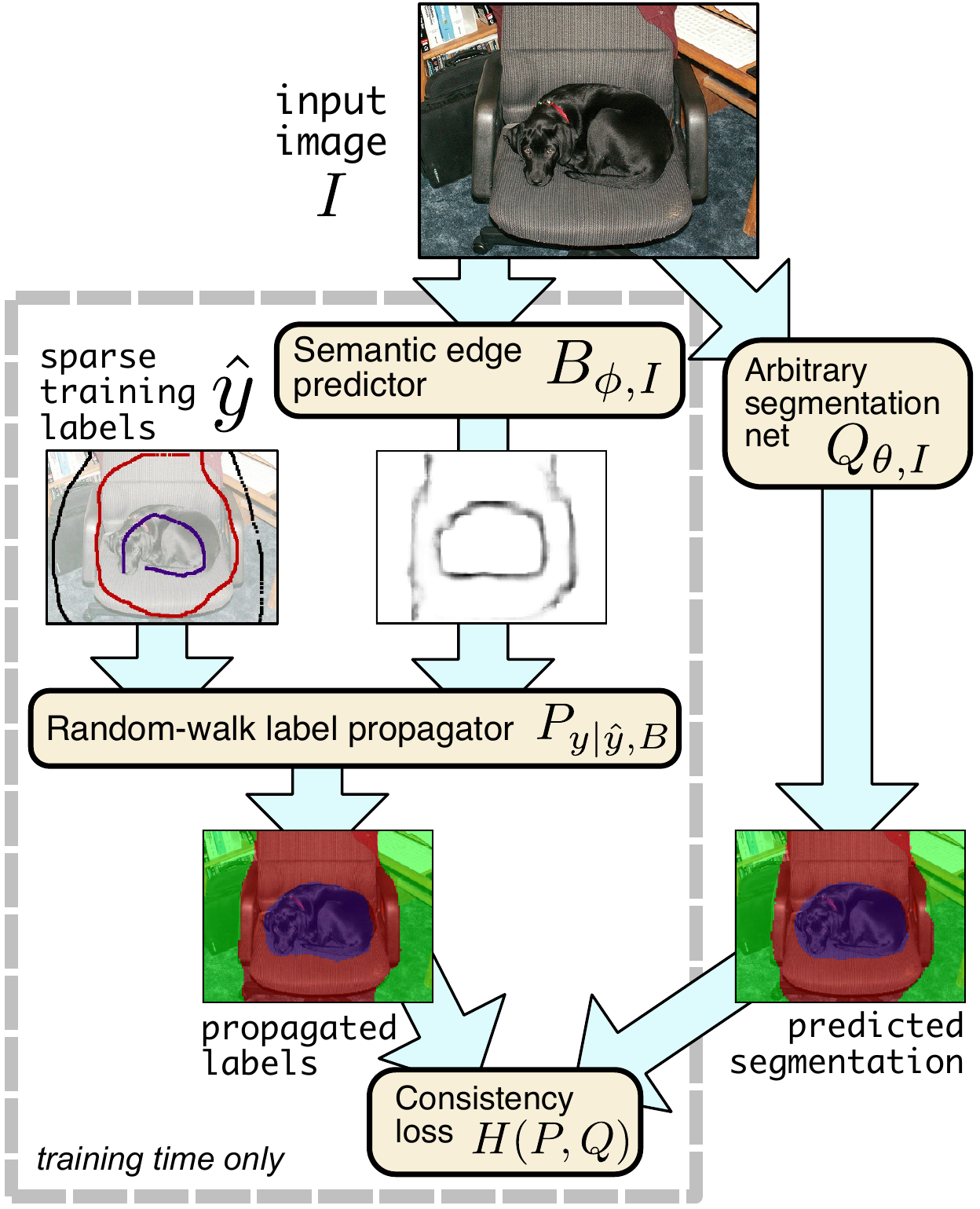}
  \caption{Overview of proposed training method.  \label{fig:blockDiagram}}
\end{figure}

We consider the task of semantic segmentation, which is to learn a
predictor capable of accurately assigning a semantic label to each
pixel in an image.  As with many other popular vision problems,
convolutional neural nets (CNNs) have emerged as the leading tool to
solve semantic segmentation problems, in part due to their ability to
leverage large datasets effectively. However, datasets for semantic
segmentation remain orders of magnitude smaller than for tasks such as
classification and detection, due chiefly to the much higher annotation
expense of this task.

To ease the annotation burden, and in line with previous
work~\cite{bearman2015s,Lin2016scribbleSup}, we propose a method for
training CNN-based semantic segmentation networks given {\em sparse}
annotations, such as the scribbles depicted in
Fig.~\ref{fig:blockDiagram}, which also depicts our proposed training
strategy.  The idea of our method is to learn mutually-consistent
networks for {\em propagating} the sparse labels to unlabeled points,
and {\em predicting} the true labeling given the image alone.
Optimizing a mutual-consistency objective obviates the need for {\em
  dense} (or, fully-labeled) supervision.  A key innovation of our
approach is proposing to use a specific, probabilistic model of sparse
label propagation that is a differentiable function of semantic
boundary predictions.  As it is differentiable, minimizing our loss
via gradient-based methods results in simultaneous learning of an
image-to-semantic-boundary predictor and an
image-to-semantic-segmentation predictor, despite having no direct
observations of semantic edges.

Our method is comparable to recent work by Lin
\etal~\cite{Lin2016scribbleSup}, which proposes alternating between
propagating sparse labels using a CRF defined over superpixels, and
training a CNN to predict the labels thus inferred.  A disadvantage of
this approach relative to ours is that~\cite{Lin2016scribbleSup}
employs a notion of label smoothness that is non-adaptive:
specifically, it is assumed that labels are constant within
superpixels, and the CRF binary potentials are not learned.  This
ultimately places an artificial upper bound on the accuracy of the
training data observed by the CNN---an upper bound that never improves
as more data is collected.  By contrast, by expressing label
propagation in terms of a learned semantic boundary predictor, we are
able to learn a concept of label propagation that is entirely
data-driven, enabling our method to scale fully with the data.
Furthermore, the probabilistic nature of our label propagation method
allows us to obtain uncertainty estimates that are directly 
incorporated into our learning process, mitigating the possibilty 
of training on propagated labels that are incorrect.

A crucial technical component of our approach is defining the label
propagation process in terms of random-walk hitting
probabilities~\cite{grady2006random}, which enables efficient
inference and gradient-based learning.  For this reason, we refer to
our approach as \ourMethod{}, a contraction of RAndom-walk
WeaKly-supervised Segmentation.

\section{Method}

\begin{table} \caption{Notation} \label{tab:notation}
  \begin{center}
    \begin{tabular}{ll}
      \toprule
      $\mX \subset \mathbb Z^2$ &  set of all pixel locations\\
      $\mhatX \subset \mX$ &  set of labeled pixels\\
      $\mL$ &  set of semantic labels\\
      $\my : \mX \rightarrow \mL$ &  a semantic labeling function\\
      $\mhaty : \mhatX \rightarrow \mL$ &  a sparse labeling\\
      $\mIm$ & a generic image\\
      $\mtheta, \mphi$ & predictor parameters\\
      $\msimp^n$ &  $n$-dim. probability simplex\\
      $\mQ_{\theta, I} : \mX \rightarrow \msimp^{\vert \mL \vert}$ & label predictor\\
      $\mB_{\phi, I} : \mX \rightarrow \mathbb R^+$ & boundary predictor\\
      $\mXi \subset \mathbb Z^+ \rightarrow \mX$ &  set of paths across image\\
      $\mxi \in \mXi$ &  a random walk\\
      $\mtau : \mXi \rightarrow \mathbb Z^+$ & path stopping time\\
      $\mP_{a \mid b} \in \msimp$ &  distribution of $a$ given $b$\\
      $\delta_i \in \{0,1\}^{\vert \delta_i \vert} \cap \msimp^{\vert \delta_i \vert}$ &  Kronecker delta vector\\
      $H(p)$ &  Shannon entropy of $p$\\
      $H(p,q)$ &  cross-entropy of $p, q$\\
      $\kld{p}{q}$ &  KL-divergence of $p, q$\\
      \bottomrule
    \end{tabular}
  \end{center}
\end{table}

Given densely labeled training images, typical approaches for
deep-learning-based semantic segmentation minimize the following
cross-entropy loss~\cite{long2015fully,chen14semantic}:
\begin{equation}
  \min_{\mtheta \in \msetTheta} \sum_{\mx \in \mX}
  H(\delta_{\my(\mx)} \,, \mQ_{\mtheta, \mIm}(\mx)),
  \label{eq:segmentationLoss}
\end{equation}
where $\delta_{y(x)} \in \msimp^{\vert \mL \vert}$ is the indicator
vector of the ground truth label $y(x)$ and $\mQ_{\mtheta, \mIm}(x)$
is a label distribution predicted by a CNN with parameters $\mtheta$
evaluated on image $\mIm$ at location $\mx$. See
Table~\ref{tab:notation} for notation. In our case, dense labels
$\my(\mx)$ are not provided: we only have labels $\mhaty$ provided at
a subset of points $\mhatX$.  Our solution is to simultaneously {\em
  infer} the dense labeling $\my$ given the sparse labeling $\mhaty$
and {\em use} the inferred labeling to train $\mQ$.  To achive this,
we propose to train our predictor to minimize the following loss:
\begin{equation}
  \min_{\mtheta, \mphi} \sum_{\mx \in \mX}
  H(\mP_{\my(\mx) \mid \mhaty, \mB_{\mphi, \mIm}} \,, \mQ_{\mtheta, \mIm}(\mx)).
  \label{eq:ourLoss}
\end{equation}
Here, $\mQ$ is a predictor of the same form as before, while
$\mP_{\my(\mx) \mid \mhaty, \mB_{\mphi, \mIm}}$ is a predicted label
distribution at $\mx$, conditioned on predicted {\em semantic
  boundaries} $\mB_{\mphi, \mIm}$ and the sparse labels.
$\mB_{\mphi, \mIm}$ is assumed to be a nonnegative-valued CNN-based
boundary predictor, in a similar vein
as~\cite{chen16edgeNet,Xie_2015_ICCV}.  See
Fig.~\ref{fig:blockDiagram} for a graphical overview of our model.

The key to our method is the definition of the propagated label
distributions $\mP_{\my \mid \mhaty, \mB_{\mphi, \mIm}}$.  As in prior
work on interactive segmentation~\cite{grady2006random}, we define
these distributions in terms of random-walk hitting probabilities,
which can be computed analytically, and which allows us to compute the
derivatives of the propagated label probabilities with respect to the
predicted boundaries.  This enables us to minimize~\eqref{eq:ourLoss}
by pure backpropagation, without tricks such as the alternating
optimization methods employed in prior weakly-supervised segmentation
methods~\cite{dai2015boxsup,Lin2016scribbleSup}.

\begin{figure}
  \centering
  \begin{subfigure}{3in}
    \includegraphics[width=3in]{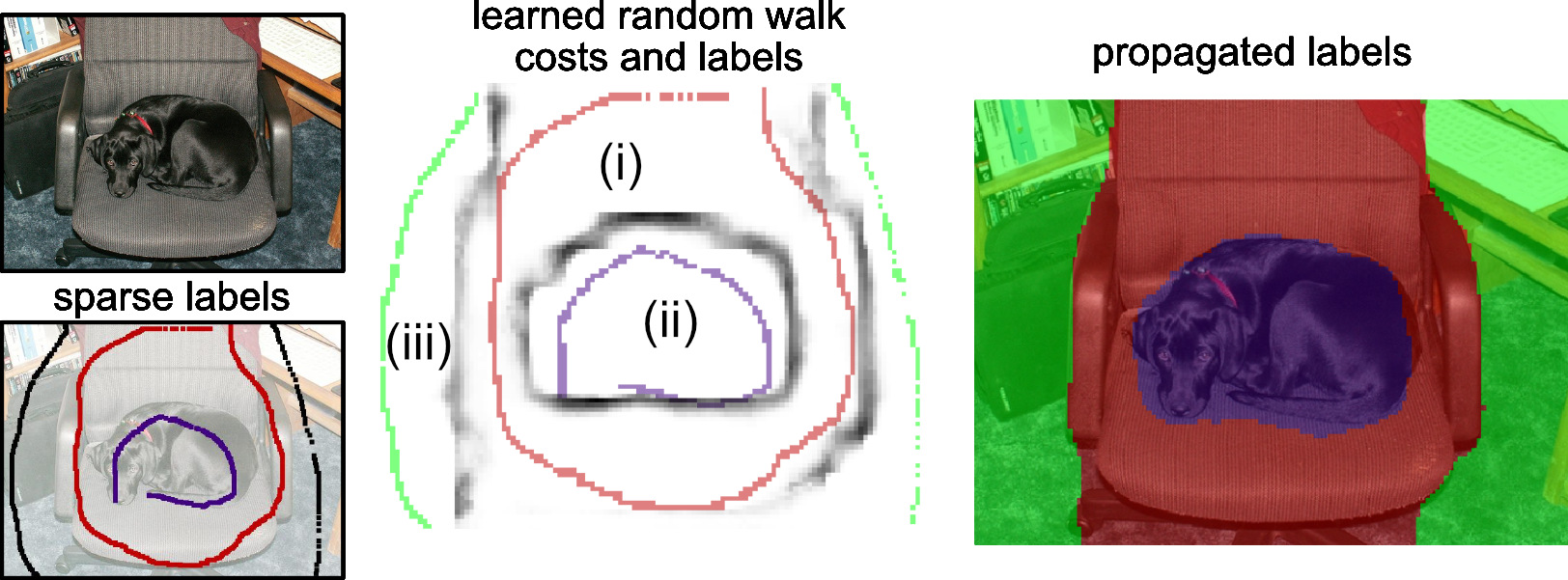}
    \caption{\label{fig:rwPropEx}}
  \end{subfigure}\\
  \begin{subfigure}{3in}
    \includegraphics[width=3in]{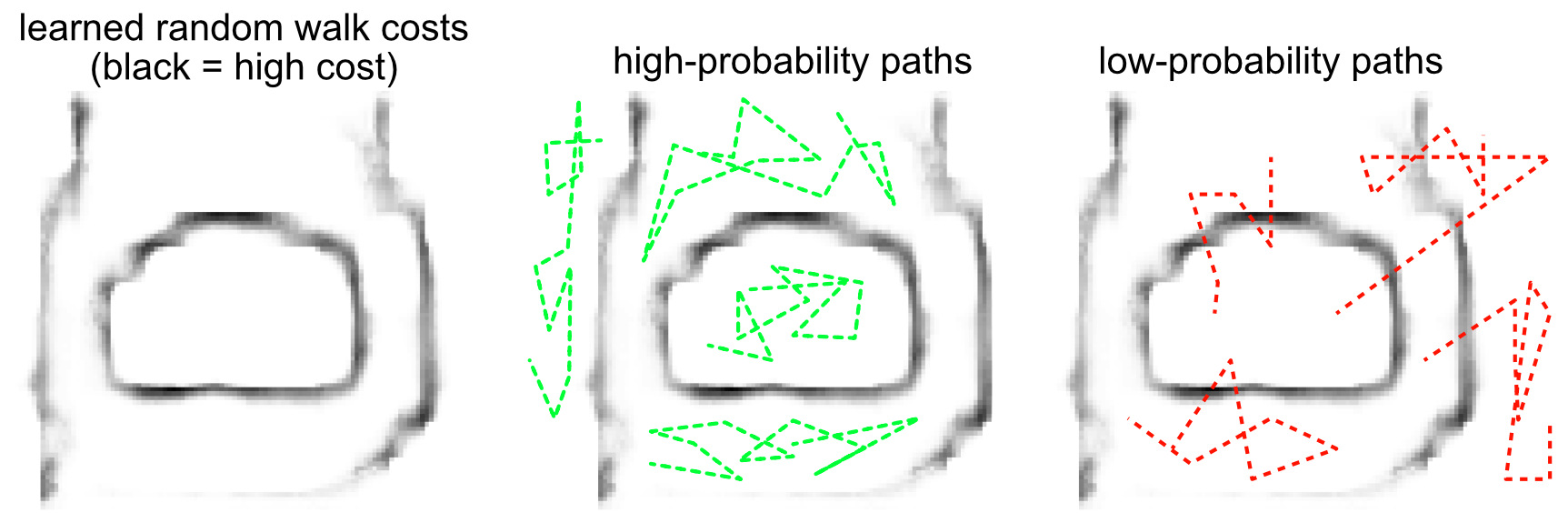}
    \caption{\label{fig:rwProbPaths}}
  \end{subfigure}
  \caption{Illustration of random-walk-based model for defining
    label probabilities based on boundaries and sparse labels.
    \label{fig:rw}}
\end{figure}

To elaborate, we first define the set of all 4-connected
paths on the image that end at the first labeled point encountered:
\begin{align}
  \mXi = \{ (\mxi_0, \mxi_1, \dots, & \mxi_{\mtau(\mxi)}) \mid \forall t,
  \mxi_t \in \mX, \Vert \mxi_{t+1} - \mxi_t \Vert_1 = 1, \notag \\ &
  \mxi_{\mtau(\mxi)} \in \mhatX, \forall t' < \mtau(\mxi) \,, \mxi_{t'}
  \notin \mhatX \}.
\end{align}
We now assign each path $\mxi \in \mXi$ a
probability that decays exponentially as it crosses boundaries:
\begin{equation}
  P(\mxi) \propto 
  \exp \left( - \sum_{t=0}^{\tau(\mxi) - 1} \mB_{\mphi, \mIm}(\mxi_t) \right)
  \frac{1}{4}^{\tau(\xi)},
  \label{eq:pathProb}\footnote{The factor $4^{-\tau(\xi)}$ is sufficient
    to ensure convergence of the partition function as long as
    $\mB_{\mphi, \mIm} \geq 0$. This factor was mistakenly omitted in the 
    original version of this paper.}
\end{equation}
where $\mB_{\mphi, \mIm}$ is assumed to be a nonnegative boundary score
prediction.  High- and low-probability paths under this model are
illustrated in Fig.~\ref{fig:rwProbPaths}.  Given sparse labels
$\mhaty$, the probability that a pixel $\mx$ has label $y'$ is then
defined as the probability that a path starting at $\mx$ eventually
hits a point labeled $y'$, given the distribution over
paths~\eqref{eq:pathProb}:
\begin{equation}
  \mP_{\my(\mx) \mid \mhaty, \mB_{\mphi, \mIm}}(y') =
  P(\{ \mxi \in \mXi \mid \mxi_0 = \mx, \mhaty(\mxi_{\tau(\mxi)}) =
  y' \} ). \label{eq:pathLabelProb}
\end{equation}
This quantity can be computed efficiently by solving a sparse linear
system, as described in Sec.~\ref{sec:rw}.  The random-walk model is
illustrated in Fig.~\ref{fig:rwPropEx}.  
Any pixel in the region labeled (ii), for example, is very likely to
be labeled {\em dog} instead of {\em chair} or {\em background},
because any path of significant probability starting in (ii) will hit
a pixel labeled {\em dog} before it hits a pixel with any other label.

To recap, the overall architecture of our method is summarized in
Fig.~\ref{fig:blockDiagram}.  At training time, an input image is
passed to an arbitrary segmentation predictor $\mQ_{\mtheta, \mIm}$
and an arbirtary semantic edge predictor $\mB_{\phi, \mIm}$.  The
semantic edge predictions and sparse training labels $\mhaty$ are
passed to a module that computes propagated label probabilities
$P_{\my\mid \mhaty, \mB}$ using the random-walk model described above.
The propagated label probabilities $P$ and the output of the predictor
$Q$ are then passed to a cross-entropy loss $H(P,Q)$, which is
minimized in the parameters $\mtheta$ and $\mphi$ via backpropagation.
At test time, $\mQ$ is evaluated and used as the prediction.
$\mP$ is not evaluated at test time, since labels are not available.

\subsection{Probabilistic justification}

\begin{figure}
  \centering
  \includegraphics[width=1.25in]{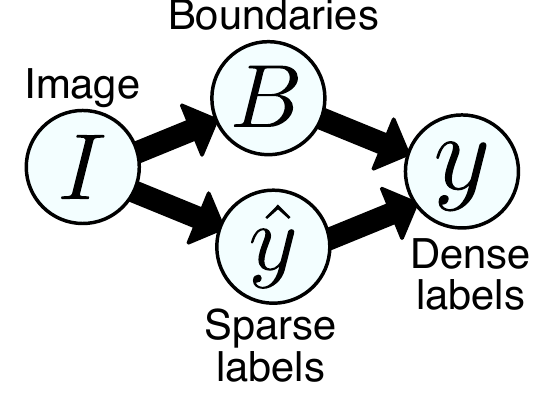}
  \caption{Independence assumptions as a graphical model 
    \label{fig:gm}}
\end{figure}
The proposed loss function~\eqref{eq:ourLoss} arises from a natural
probabilistic extension of~\eqref{eq:segmentationLoss} to the case
where dense labels are unobserved.  Specifically, we consider
marginalizing~\eqref{eq:segmentationLoss} over the unobserved dense
labels, given the observed sparse labels.  This requires us to define
$P(\my \mid \mhaty, \mIm)$.  We submit that a natural way to do so is
to introduce a new variable $B$ representing the image's semantic
boundaries.  This results in the proposed graphical model in
Fig.~\ref{fig:gm} to represent the independence structure of
$\my,\mhaty,\mB,\mIm$.

It is then straightforward to show that~\eqref{eq:ourLoss} is
equivalent to marginalizing~\eqref{eq:segmentationLoss} with respect to a
certain distribution $P(\my \mid \mhaty, \mIm)$, after making a few
assumptions.  First, the conditional independence structure depicted
in Fig.~\ref{fig:gm} is assumed.  $y(x)$ and $y(x')$ are assumed
conditionally independent given $\mhaty, \mB,$
$\forall \mx \neq \mx' \in X$, which allows us to specify
$P(\my \mid \mhaty, \mB)$ in terms of marginal distributions and
simplifies inference.  Finally, $\mB$ is assumed to be a deterministic
function of $\mIm$, defined via parameters $\mphi$.  

We note that the conditional independence assumptions made in
Fig.~\ref{fig:gm} are significant.  In particular, $\my$ is assumed
independent of $\mIm$ given $\mhaty$ and $\mB$.  This essentially
implies that there is at least one label for each connected component
of the true label image, since knowing the underlying image usually
does give us information as to the label of unlabeled connected
components.  In practice, strictly labeling every connected component
is not necessary.  However, training data that egregiously violates
this assumption will likely yield poor results.

\subsection{Random-walk inference} \label{sec:rw}

Key to our method is the efficient computation of the random-walk
hitting probabilities~\eqref{eq:pathLabelProb}.  It is well-known that
such probabilities can be computed efficiently via solving linear
systems~\cite{grady2006random}.  We briefly review this result here.

The basic strategy is to compute the {\em partition function}
$Z_{xl}$, which sums the right-hand-side of~\eqref{eq:pathProb} over
all paths starting at $x$ and ending in a point labeled $l$.  We can
then derive a dynamic programming recursion expressing $Z_{xl}$ in
terms of the same quantity at neighboring points $x'$.  This 
recursion defines a set of sparse linear constraints on $Z$, which 
we can then solve using standard sparse solvers.

We first define
$\mXi_{xl} := \{\mxi \in \mXi \mid \mxi_0 = \mx, \,
\mhaty(\mxi_{\mtau(\mxi)}) = l\}$. $Z_{xl}$ is then defined as
\begin{equation}
  Z_{\mx l} := \sum_{\mxi \in \mXi_{\mx l}} 
  \exp \left(- \sum_{t=0}^{\mtau(\mxi)-1} \mB_{\mphi,\mIm}(\mxi_t) \right)
  \frac{1}{4}^{\tau(\xi)}.
\end{equation}
The first term in the inner sum can be factored out by introducing a
new summation over the four nearest neighbors of $\mx$, denoted
$x' \sim \mx$, easily yielding the recursion
\begin{align}
  Z_{\mx l} & = 
  \frac{1}{4} e^{- \mB_{\mphi,\mIm}(\mx)}
  \sum_{\substack{x' \sim \mx, \\ \mxi \in \mXi_{x'l}}}
  \frac{1}{4}^{\tau(\xi)} e^{- \sum_{t=0}^{\mtau(\mxi)-1} \mB_{\mphi,\mIm}(\mxi_t)} \notag\\
  & = \frac{1}{4} e^{- \mB_{\mphi,\mIm}(\mx)}
    \sum_{x' \sim \mx} Z_{x'l}. \label{eq:zRec}
\end{align}
Boundary conditions must also be considered in order to fully constrain
the solution.  Paths exiting the image are assumed to have zero
probability: hence, $Z_{\mx l} := 0, \, \forall \mx \notin \mX$.
Paths starting at a labeled point $\mx \in \mhatX$ immediately
terminate with probability 1; hence,
$Z_{\mx l} := 1, \, \forall \mx \in \mhatX$.  Solving this 
system yields a unique solution for $Z$, from which the desired 
probabilities are computed as
\begin{equation}
  \mP_{\my(\mx) \mid \mhaty, \mB_{\mphi, \mIm}}(y') = 
  \frac{Z_{\mx y'}}{\sum_{l \in \mL} Z_{\mx l}}.
\end{equation}

\subsection{Random-walk backpropagation} \label{sec:rwDeriv}

In order to apply backpropagation, we must ultimately compute the
derivative of the loss with respect to a change in the boundary score
prediction $B_{\mphi,\mIm}$.  Here, we focus on computing the
derivative of the partition function $Z$ with respect to the boundary
score $B$, the other steps being trivial.

Since computing $Z$ amounts to solving a linear system, this turns out
to be fairly simple. Let us write the constraints~\eqref{eq:zRec} in
matrix form $A z = b$, such that $A$ is square, $z_i := Z_{\mx_i}$
(assigning a unique linear index $i$ to each $\mx_i \in \mX$, and
temporarily omitting the dependence on $l$), and the $i$th rows of
$A,b$ correspond to the constraints
$C_i Z_{\mx_i} = \sum_{x' \sim \mx_i} Z_{x'}$, $Z_{\mx_i} = 0$, or
$Z_{\mx_i} = 1$ (as appropriate), where
$C_i := \exp (\mB_{\mphi,\mIm}(\mx_i) )$.  Let us consider the effect
of adding a small variation $\epsilon V$ to $A$, and then re-solving
the system.  It can be shown that
\begin{equation}
  (A+\epsilon V)^{-1} b = z - \epsilon A^{-1} V z + O(\epsilon^2).
\end{equation}
Substituting the first-order dependence on $V$ into a Taylor 
expansion of the loss $L$ yields:
\begin{align}
  L((A + \epsilon V)^{-1} b) 
  & = L(z) - 
    \left< \dod{L}{z}, \epsilon A^{-1} V z  \right>
    + O(\epsilon^2) \notag\\
  & = L(z) - 
    \left< A^{-1 \intercal} \dod{L}{z} z^\intercal, \epsilon V \right>
    + O(\epsilon^2) \notag.
\end{align}
A first-order variation of $C_i$ corresponds to $V_i = -\delta_{ii}$,
which implies that 
\begin{equation}
  \dod{L}{C_i} = \left( A^{\intercal -1} \dod{L}{z} \right)_i z_i.
\end{equation}
In summary, this implies that computing the loss derivatives with
respect to the boundary score can be implemented efficiently by
solving the sparse adjoint system
$A^{\intercal}\od{L}{C} = \od{L}{z}$, and multiplying the result
pointwise by the partition function $z$, which in turn allows us to
efficiently incorporate sparse label propagation as a function of
boundary prediction into an arbitrary deep-learning framework.

\subsection{Uncertainty-weighting the loss} \label{sec:weights}

\begin{figure}
  \centering
  \includegraphics[width=2.5in]{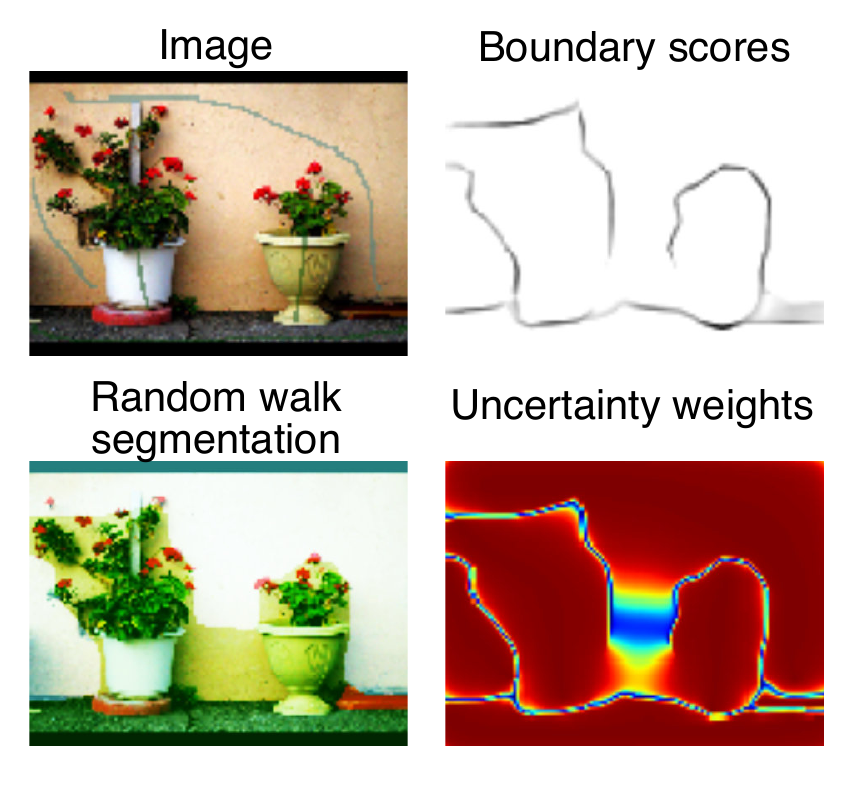}
  \caption{Visualization of loss uncertainty weights (blue = low
    weight, red = high weight) \label{fig:lossWeights}}
\end{figure}

An advantage of our method over prior work is that the random-walk
method produces a distribution over dense labelings
$\mP_{\my \mid \mhaty, \mB_{\mphi, \mIm}}$ given sparse labels, as
opposed to a MAP estimate.  These uncertainty estimtates can be used
to down-weight the loss in areas where the inferred labels may 
be incorrect, as illustrated in Fig.~\ref{fig:lossWeights}.
In this example, the boundary predictor failed to correctly 
predict parts of object boundaries.  In the vicinity of 
these gaps, the label distribution is uncertain, and the 
MAP estimate is incorrect.  However, we can mitigate the 
problem by down-weighting the loss proportional to the 
uncertainty estimate.

More concretely, we actually minimize the following modification of
the loss~\eqref{eq:ourLoss}:
\begin{align}
  \sum_{\mx \in \mX}
  w(x) \kld{\mP_{\my(\mx) \mid \mhaty, \mB_{\mphi, \mIm}}}
  {\mQ_{\mtheta, \mIm}(\mx)}
  + H(\mP_{\my(\mx) \mid \mhaty, \mB_{\mphi, \mIm}}), \label{eq:loss}
\end{align}
where we define
$w(x) := \exp (- \alpha H(\mP_{\my(\mx) \mid \mhaty, \mB_{\mphi,
    \mIm}}))$,
for some fixed parameter $\alpha$. This loss reduces
to~\eqref{eq:ourLoss} for the case $w(x) = 1$.  Although the KL
component of the loss can be avoided by increasing the prediction
entropy, the explicit entropy regularization term prevents 
trivial solutions of very large entropy everywhere.

\section{Related work}

The method most comparable to ours is the work of Lin
\etal~\cite{Lin2016scribbleSup}.  In contrast
to~\cite{Lin2016scribbleSup}, our method features
fully-differentiable, gradient-based training (as opposed to
alternating optimization); we learn an inductive rule for predicting
boundaries and propagating labels, as opposed to using non-adaptive
superpixels and a CRF with non-adaptive binary potentials, which
enables us to adapt to large datasets in a data-driven way; and we
employ a probabilistic notion of label propagation that enables us to
define an uncertainty-weighted loss that mitigates the possibilty of
training on propagated labels that are incorrect.  Another notable
method in the same vein as~\cite{Lin2016scribbleSup} is the BoxSup
method~\cite{dai2015boxsup}, which also employs alternating
optimization, but uses bounding-box annotations as weak supervision.

Our method was initially inspired by~\cite{bearman2015s}, which
introduced the idea of training on what we refer to here as {\em
  sparse} labels as a source of weak supervision.  Instead of
attempting to directly propagate labels, as we do, that method
leverages a notion of objectness to mitigate overfitting.  A few other
works have proposed different modes of weak supervision for
segmentation.  Notably,~\cite{xu2015learning}
and~\cite{pathak2015constrained} both model weak supervision as
imposing linear constraints to be satisfied by the predictor,
resulting in models trained by alternating optimization.  Our method
can be viewed from a similar perspective, since we also impose our
weak supervision via linear constraints~\eqref{eq:zRec}; however, our
constraints explicitly model the process of spatial label propagation,
whereas the constraints proposed
in~\cite{xu2015learning,pathak2015constrained} model only aggregate
statistics over regions, and hence make no provision for learning
boundaries as we do in this work.  Furthermore, our model is
differentiable and can be optimized via gradient-based methods.

To the extent that it learns semantic edges with a CNN, our method is
similar to previous works such as~\cite{Xie_2015_ICCV}, which
also learns semantic edges using a CNN.  However,~\cite{Xie_2015_ICCV}
achieves this using direct supervision of edges---which we do not
require---and does not jointly train a semantic labeler, as we do.
Learning edges with a weaker form of edge supervision is proposed
in~\cite{Khoreva2016Cvpr}; however, this method relies on 
a combination of heuristic boundary detectors and bounding boxes
for supervision.  To reiterate, our method works without any 
heuristic source of boundaries as input.

Another vein of prior work relates to some form of joint reasoning
about boundaries and segments.  Most prominently, random walks were
previously applied to interactive segmentation
in~\cite{grady2006random}.  However, that work did not consider
{\em learning} these random walks by learning boundary scores, 
as we do, nor did it consider the task of semantic image 
segmentation in general.  More recently,~\cite{chen16edgeNet}
proposed a method to jointly learn semantic edges and a CNN-based
semantic labeler in an gradient-based learning framework.
However, their method applies only to the strong-supervision case,
and cannot leverage sparse annotations of the kind we employ here.


\section{Experiments}

We implemented \ourMethod{} in the Caffe~\cite{jia2014caffe}
framework. The semantic-boundary-prediction network $\mB_{\mphi,\mIm}$
and the semantic-segmentation network $\mQ_{\mtheta,\mIm}$ were both
implemented as fully-convolutional CNNs, based on the same
ResNet-101~\cite{he2016resnet} architecture.  The final
average-pooling and fully-connected layers were removed, and features
from the last resulting layer were upsampled and combined with
intermediate-layer features to produce a 4x-downsampled output for
both the semantic boundary and label predictions.  Both networks were
initialized from a model trained for classification on the ImageNet
2012 dataset.  We applied no data augmentation techniques in training
any of the methods.

\ourMethod{} was trained on the publicly available scribble
annotations provided by~\cite{Lin2016scribbleSup}.  We used the same
training and validation splits as~\cite{Lin2016scribbleSup}, for both
the PASCAL VOC 2012 and PASCAL CONTEXT datasets: for VOC, the
validation set consisted of the VOC 2012 validation set, while the
training set consisted of all other images labeled in either the VOC
2012 dataset or the PASCAL Semantic Boundary
dataset~\cite{hariharan2011semantic} (10582 training images, and 1449
validation images). We trained models for each of these datasets
independently.

We evaluated the performance of both the semantic-segmentation network
$Q$ and the label-propagation network $P$ (propagating the sparse
labels given the learned boundaries).  To summarize, evaluating the
predicted labels $Q$ on the validation set, \ourMethod{} slightly
underperformed the published results of~\cite{Lin2016scribbleSup} on
VOC 2012, while slightly outperforming~\cite{Lin2016scribbleSup} on
CONTEXT.  Our other major observation was that the propagated labels
$P$ on the training set were approximately as accurate as the {\em
  best possible} labeling of a superpixel segmentation of the images.

To elaborate, in Table~\ref{tab:voc}, {\em MIOU} refers to the
mean-intersection-over-union metric, while {\em w/ CRF} refers to the
same metric evaluated after post-processing the results with a
fully-connected CRF, as in~\cite{Lin2016scribbleSup}.  {\em
  \ourMethod{} $Q_{\mtheta, \mIm}$} refers to the evaluation of the
predicted labels $Q$ on the validation set given the image alone,
after jointly training $P$ and $Q$ via SGD. {\em \ourMethod{} train
  P,Q then Q} also refers to evaluation of $Q$, but with a slightly
different training protocol: in this case after jointly training $P$
and $Q$ with SGD, $Q$ was fine-tuned with $P$ fixed.  {\em
  \ourMethod{} training $P_{\my \mid \mhaty, B}$, 0\% abstain} refers
to evaluation of $P$ on the {\em training set}, given the sparse
training labels and learned boundaries $\mB_{\mphi, \mIm}$.  {\em
  \ourMethod{} training $P_{\my \mid \mhaty, B}$, 6\% abstain}
consists of the same evaluation, but allowing $P$ to abstain from
prediction on 6\% of the pixels, which (for this particular model)
corresponds to abstaining on all pixels with a confidence score $w(x)$
below 0.5 (c.f. Sec.~\ref{sec:weights}).
The next section of Table~\ref{tab:voc} reports baselines: {\em
  sparse-loss baseline} consists of training the same base network
$Q$, but using loss~\eqref{eq:segmentationLoss} (i.e., without $P$),
evaluating it only at the sparse locations $\mhatX$. {\em train on
  dense ground truth} is the result we obtain training our base
network $Q$ on the dense ground-truth training data.  {\em
  ScribbleSup} refers to the result reported
by~\cite{Lin2016scribbleSup}, which we report here verbatim.  We note
that~\cite{Lin2016scribbleSup} used a different base segmentation
network (DeepLab) than we used in our experiments.  

The last section of Table~\ref{tab:voc} reports statistics for
baselines meant to represent best-case performance bounds for a
superpixel-based method such as~\cite{Lin2016scribbleSup}.  These were
obtained by segmenting the input images using the
method~\cite{felzenszwalb2004efficient} (with author-suggested
parameters), and labeling the resulting superpixels in different ways.
SPOPT corresponds to labeling each superpixel with the majority label
from the ground-truth dense segmentation.  SPCON differs from SPOPT
only on superpixels containing scribble annotations: to these, SPCON
assigns the majority label from the scribbles contained within.  {\em
  Train on SPOPT/SPCON} refer to training predictors $Q$ using the
labelings of SPOPT and SPCON, respectively.  These results are
interesting for a number of reasons.  First, the 
propagated labelings $P$ that we deduce in the course of training, are
nearly as good (VOC) or better (CONTEXT) than the best
possible results obtainable using superpixels.  Second, we see that
{\em training with} our propagated labelings $P$ is competitive with
training on optimal superpixel labelings.  Finally, we emphasize that
while the superpixel baselines cannot improve with training data (as
they are not trained), our label propagation model is naturally
refined as we train on larger datasets.

In relative terms, \ourMethod{} performed better on the CONTEXT
dataset, as evidenced in Table~\ref{tab:context}.  One potential
reason for this is the greater number of classes for this dataset (60
vs. 21 for VOC), which naturally calls for finer boundaries.  Since
our method is able to adaptively learn boundaries suited to the task,
while~\cite{Lin2016scribbleSup} uses non-adaptive heuristics to
generate superpixels, this may account for the better relative
performance of \ourMethod{} in this context.  We also hypothesize that
it is easier to learn semanic boundaries when there are a greater
number of classes, because low-level edges and features become a more
informative cue in this case.  Surprisingly, our dense ground-truth
baseline performed significantly worse than \ourMethod{}; we
hypothesize this is due to overfitting, a consequence of the smaller
amount of training data and increased number of classes in CONTEXT,
exacerbated by our use of the very-deep ResNet model.  Joint training
of the propagator network $P$ in \ourMethod{} seems to have a
regularizing effect that may have prevented overfitting to some
extent.

\begin{figure*}
  \centering
  \includegraphics[width=6in]{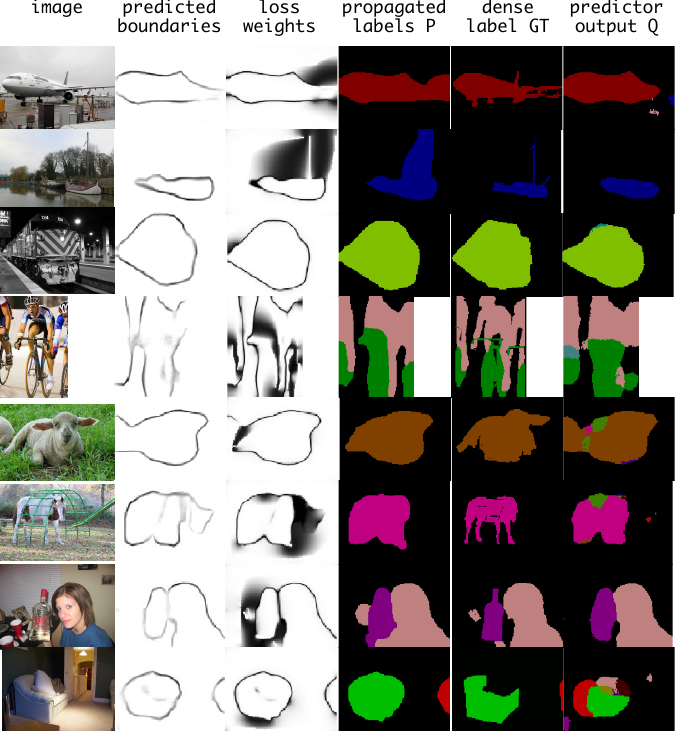}
  \caption{Validation set results on VOC 2012 dataset
    \label{fig:voc}}
\end{figure*}

\begin{figure*}
  \centering
  \includegraphics[width=6in]{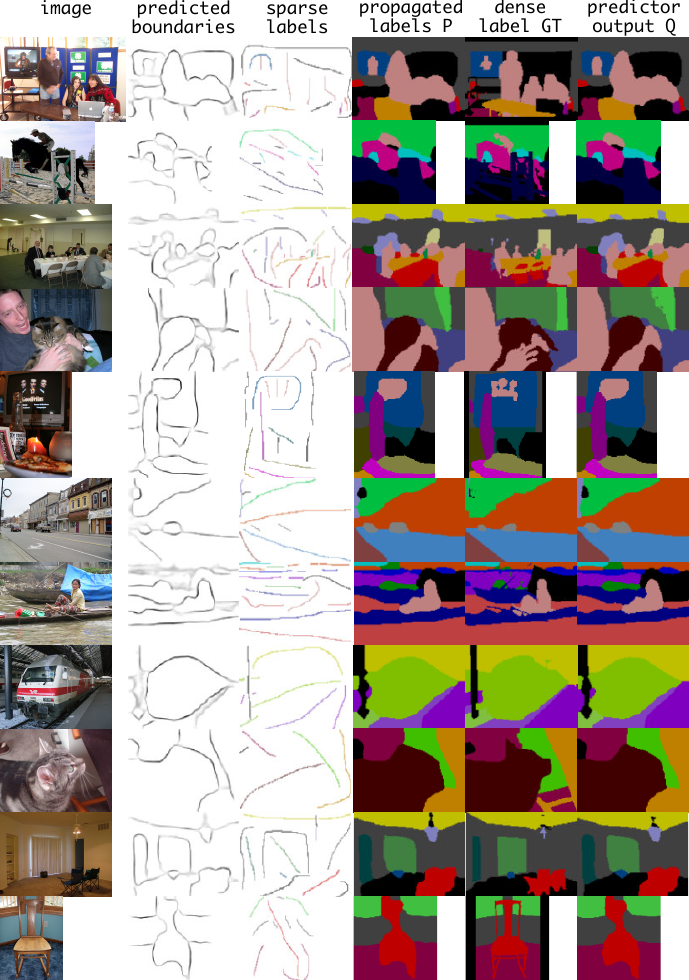}
  \caption{Training results on CONTEXT dataset \label{fig:context}}
\end{figure*}

Qualitative validation-set results are shown in Fig.~\ref{fig:voc} for
the VOC 2012 dataset, while CONTEXT training-set results are shown in
Fig.~\ref{fig:context}.  The training-set results of
Fig.~\ref{fig:context} demonstrate that \ourMethod{} is able to deduce
high-quality semantic boundaries, thereby producing propagated
labelings $P$ that are a close approximation to the ground truth dense
labelings (which are not used at training time, to be clear).  In the
validation-set results of Fig.~\ref{fig:voc}, the loss weights $w(x)$
and propagated labels $P$ are shown in addition to the predictions
$Q$---to be clear, these depend on the sparse labels for these
specific examples, which were not used to train this model.  Here we
remark that our semantic boundary predictions also generalize well to
the validation set.  Although we did not train on these images, we
also observe that had we done so, the loss weights would have behaved
appropriately, down-weighting the loss in regions where the propagated
labels are incorrect.  This seems to happen most often in regions with
very fine boundaries (such as the mast of the boat and the airplane's
wing), where our limited resolution sometimes causes missed
boundaries.
 
In general, we note that subjectively, resolution seemed to be a
limiting factor in the accuracy of our boundary prediction and label
propagation steps.  We used quarter-resolution outputs (typicaly
around 128x96 pixels) for these steps in order to minimize the
computational cost of computing random-walk hitting probabilities. An
average forward-backwards pass of the entire network took about 1.1 s
per image, with about 800 ms of that spent solving for the random-walk
hitting probabilities.  This layer was implemented using a CPU-based
sparse linear system solver, whereas the rest of the network was run
on the GPU (an NVIDIA GTX 1080).  We anticipate that implementing this
layer using GPU operations will allow us to increase the resolution of
these critical steps, which will in turn lead to increased prediction
accuracy.

\begin{table}
  \centering
  \begin{tabular}{lll}
    Method & MIOU & w/ CRF\\\hline\hline
    \ourMethod{} $Q_{\mtheta, \mIm}$ &  57.1 & 60.0 \\
    \ourMethod{} train $P,Q$, then $Q$ & 59.5 & 61.4 \\
    \ourMethod{} training $P_{\my \mid \mhaty, B}$, 0\% abstain & 75.8 & . \\
    \ourMethod{} training $P_{\my \mid \mhaty, B}$, 6\% abstain & 81.2 & . \\\hline
    Sparse-loss baseline & 55.8 & \\
    Train on dense ground truth & 66.3 & 68.8 \\
    ScribbleSup (reported) & . & 63.1\\\hline
    Opt. superpixel labels (SPOPT) & 83.1 & . \\
    Opt. consistent s-p. labels (SPCON) & 76.5 & . \\
    Train on SPOPT & 62.8 & .\\
    Train on SPCON & 61.1 & .\\
  \end{tabular}
  \caption{Results on VOC 2012 validation set}
 \label{tab:voc}
\end{table}

\begin{table}
  \centering
  \begin{tabular}{lll}
    Method & MIOU & w/ CRF\\\hline\hline
    \ourMethod{} $Q_{\mtheta, \mIm}$ & 36.0 & 37.4 \\
    \ourMethod{} training $P_{\my \mid \mhaty, B}$, 0\% abstain & 75.5 & . \\\hline
    Sparse-loss baseline & 26.6 & . \\
    Train on dense ground truth & 31.7 & 32.4 \\
    ScribbleSup (reported) & . & 36.1 \\\hline
    Opt. consistent s-p. labels (SPCON) & 70.2 & .
  \end{tabular}
  \caption{Results on CONTEXT validation set}
 \label{tab:context}
\end{table}

\section{Conclusions}

We have presented a novel approach to mitigating the expense of procuring labeled data in semantic segmentation, through a framework that utilizes only sparse clicks or scribbles for training. This has a significant impact on the possibilities for semantic segmentation---for a given dataset, one may obtain competitive labels at a fraction of the cost and conversely, for a given budget, one may obtain labeled data at a much larger scale. Our main technical contribution is a random-walk based label propagation mechanism, which is shown to be differentiable and usable in powerful deep neural network architectures for semantic segmentation. We achieve this through a novel predictor-propagator paradigm, which produces uncertainty estimates for inferred dense labels given sparse labels. We demonstrate encouraging results on challenging benchmarks. More importantly, we argue that our framework has inherent advantages over prior works, since our label propagation is not artificially upper-bounded by superpixel baselines, rather, can keep improving with larger-scale training data. Also, we note that our contribution is equally valid for any state-of-the-art CNN-based semantic segmentation engines. In future work, we will explore other state-of-the-art segmentation architectures and incorporate other forms of weak supervision such as bounding boxes.

{\small
\bibliographystyle{ieee}
\bibliography{rawwks}
}

\end{document}


\title{Learning random-walk label propagation \\for weakly-supervised semantic segmentation:\\Supplemental material}

\author{Paul Vernaza \qquad Manmohan Chandraker\\
NEC Labs America, Media Analytics Department\\
10080 N Wolfe Road, Cupertino, CA 95014\\
{\tt\small \{pvernaza,manu\}@nec-labs.com}
} 

\maketitle

\section{Additional results}

\subsection{Additional qualitative results}

An expanded version of the VOC 2012 result figure is shown
in Fig.~\ref{fig:vocLarge}.

\begin{figure*}
  \centering
  \includegraphics[width=6in]{figures/vocValLarge}
  \caption{Validation set results on VOC 2012 dataset
    \label{fig:vocLarge}}
\end{figure*}

\section{Derivations}

\subsection{Probabilistic justification of loss}

In Sec. 2.1, it is claimed that Eq. 2 follows from marginalizing
Eq. 1 over the unobserved labels, given certain assumptions.
This is proved here.

Marginalizing the objective in Eq. 1 over $y$ is expressed as:
\begin{align}
  \sum_{\tilde \my, \tilde \mB}
  P_{\my, \mB \mid \hat \my, \mIm}(\tilde \my, \tilde \mB)
  \sum_{\mx \in \mX}
  H(\delta_{\tilde \my(\mx)} \,, \mQ_{\mtheta, \mIm}(\mx)).
\end{align}
Using the conditional independence assumptions
and the assumption that $\mB$ is a deterministic function
of $\mIm$ yields
\begin{align}
  \sum_{\tilde \my}
  P_{\my \mid \hat \my, \mB_\mIm}(\tilde \my)
  \sum_{\mx \in \mX}
  H(\delta_{\tilde \my(\mx)} \,, \mQ_{\mtheta, \mIm}(\mx)),
\end{align}
where $\mB_\mIm$ denotes the boundaries as a function of $\mIm$.
Assuming that $y(x)$ and $y(x')$ are conditionally independent 
given $\mhaty, \mB$, $\forall \mx \neq \mx' \in \mX$ allows us 
to represent $P_{\my \mid \mhaty, \mB_{\mIm}}$ as a product 
of factors:
\begin{align}
  \sum_{\tilde \my}
  \left(
  \prod_{\mx' \in \mX} P_{\my(\mx') \mid \hat \my, \mB_\mIm}(\tilde \my(\mx'))
  \right)
  \sum_{\mx \in \mX}
  H(\delta_{\tilde \my(\mx)} \,, \mQ_{\mtheta, \mIm}(\mx)).
\end{align}
Factorizing out sums equal to one results in
\begin{align}
  \sum_{\mx \in \mX} \sum_{\tilde y(x)}
  P_{\my(\mx) \mid \mhaty, \mB_{\mIm}}(\tilde \my(\mx))
  H(\delta_{\tilde \my(\mx)} \,, \mQ_{\mtheta, \mIm}(\mx)).
\end{align}
Expanding the definition of cross-entropy yields the desired
expression:
\begin{align}
  \sum_{\mx \in \mX} 
  H(P_{\my(\mx) \mid \mhaty, \mB_\mIm}, \, \mQ_{\mtheta, \mIm}(\mx))
\end{align}

\subsection{Fr\'{e}chet derivative of matrix inverse}

This result is used in the derivation of the derivative of 
the random-walk partition function.  An intuitive derivation
is provided here.

Suppose $Ax=b$. We wish to find an expansion linear in $\epsilon V$
for $\tilde x := (A + \epsilon V)^{-1}b$, assuming the inverse exists.
\begin{align}
  (A + \epsilon V) \tilde x & = b\\
  A \tilde x & =  b - \epsilon V \tilde x\\
  \tilde x & = A^{-1}b - \epsilon A^{-1} V \tilde x\\
& = A^{-1}b - \epsilon A^{-1} V (A^{-1} b - O(\epsilon))\\
& = A^{-1}b - \epsilon A^{-1} V A^{-1} b + O(\epsilon^2)
\end{align}
where the second-to-last line follows from recursive expansion of the
same expression.



\title{Learning random-walk label propagation \\for weakly-supervised semantic segmentation:\\Supplemental material}

\author{Paul Vernaza \qquad Manmohan Chandraker\\
NEC Labs America, Media Analytics Department\\
10080 N Wolfe Road, Cupertino, CA 95014\\
{\tt\small \{pvernaza,manu\}@nec-labs.com}
} 

\maketitle

\section{Errata}

\begin{itemize}
\item In Sec. 2.1, line 275 should read ``It is then straightforward 
to show that (2) is equivalent to marginalizing (1) \ldots''.
\end{itemize}

\subsection{Random-walk backpropagation}

There are minor sign errors in Section 2.3.  Equation 9 should read:
\begin{equation}
  (A+\epsilon V)^{-1} b = z - \epsilon A^{-1} V z + O(\epsilon^2).
\end{equation}
The equation after Eq. 9 should read:
\begin{align}
  L((A + \epsilon V)^{-1} b) 
  & = L(z) - 
    \left< \dod{L}{z}, \epsilon A^{-1} V z  \right>
    + O(\epsilon^2) \notag\\
  & = L(z) - 
    \left< A^{-1 \intercal} \dod{L}{z} z^\intercal, \epsilon V \right>
    + O(\epsilon^2) \notag.
\end{align}
The next line should read: ``A first-order variation of $C_i$
corresponds to $V_i = -\delta_{ii}$.''

\section{Additional results}

\subsection{Additional quantitative results for CONTEXT}

Table~\ref{tab:context} shows additional results for the CONTEXT
dataset.  In particular, results are added for training on the
sparse-loss baseline as well as for the optimal labeling of
superpixels consistent with the scribble annotations.  As expected,
the model trained with the sparse loss (i.e., Eq (1) evaluated only at
the labeled locations) exhibited severe overfitting.  Also in line
with our expectations, we observed that the accuracy of the optimal
labeling of superpixels consistent with the scribble annotations
(SPCON) decreased relative to the VOC 2012 dataset.  We hypothesize
this is due partly to the fact that the finer classes of CONTEXT
require finer boundaries (we used the same superpixel segmentation
parameters as we used for VOC 2012).  We also note that the
training-set segmentations inferred by \ourMethod{} (the row labeled
{\em \ourMethod{} training $P_{\my \mid \mhaty, \mB}$, 0\% abstain}
were more accurate than those of SPCON ($75.5\%$ MIOU vs. $70.2\%$).
This constitutes strong evidence for the necessity of learning
boundaries (as in \ourMethod{}) as opposed to using superpixel
segmentations for propagating sparse labels.

\begin{table}
  \centering
  \begin{tabular}{lll}
    Method & MIOU & w/ CRF\\\hline\hline
    \ourMethod{} $Q_{\mtheta, \mIm}$ & 36.1 & 37.4 \\
    \ourMethod{} training $P_{\my \mid \mhaty, B}$, 0\% abstain & 75.5 & . \\\hline
    Sparse-loss baseline & 26.6 & . \\
    Train on dense ground truth & 31.7 & 32.4 \\
    ScribbleSup (reported) & . & 36.1 \\\hline
    Opt. consistent s-p. labels (SPCON) & 70.2 & .
  \end{tabular}
  \caption{Results on CONTEXT validation set}
 \label{tab:context}
\end{table}

\subsection{Additional qualitative results}

\begin{figure*}
  \centering
  \includegraphics[width=6in]{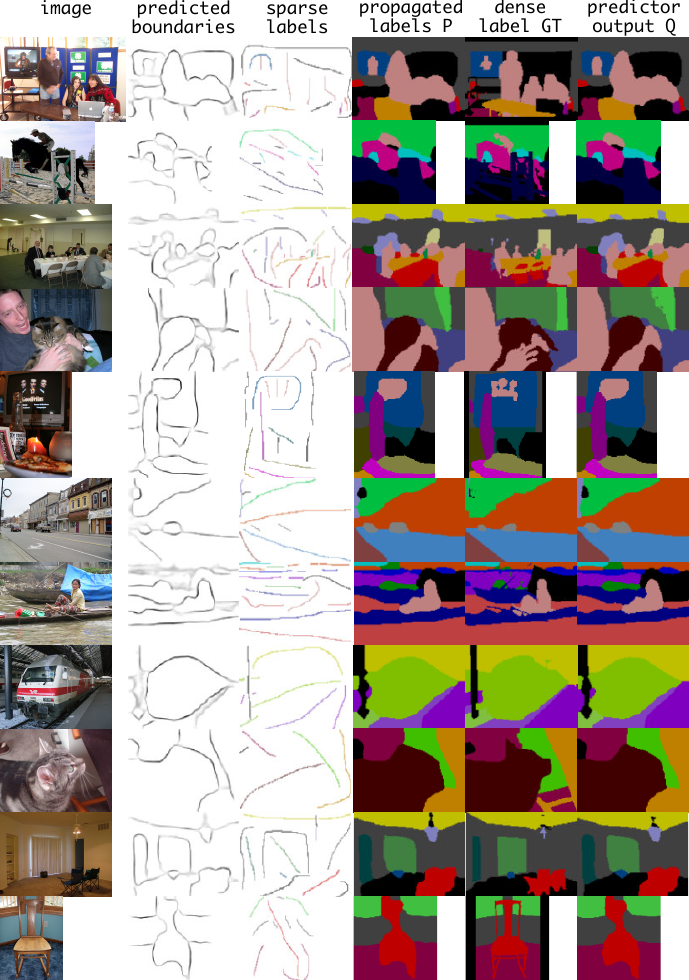}
  \caption{Training results on CONTEXT dataset \label{fig:contextLarge}}
\end{figure*}

Expanded versions of the qualitative results figures are given in
Fig.~\ref{fig:vocLarge} and Fig.~\ref{fig:contextLarge}.

\begin{figure*}
  \centering
  \includegraphics[width=6in]{figures/vocValLarge}
  \caption{Validation set results on VOC 2012 dataset
    \label{fig:vocLarge}}
\end{figure*}

\section{Derivations}

\subsection{Probabilistic justification of loss}

In Sec. 2.1, it is claimed that Eq. 2 follows from marginalizing
Eq. 1 over the unobserved labels, given certain assumptions.
This is proved here.

Marginalizing the objective in Eq. 1 over $y$ is expressed as:
\begin{align}
  \sum_{\tilde \my, \tilde \mB}
  P_{\my, \mB \mid \hat \my, \mIm}(\tilde \my, \tilde \mB)
  \sum_{\mx \in \mX}
  H(\delta_{\tilde \my(\mx)} \,, \mQ_{\mtheta, \mIm}(\mx)).
\end{align}
Using the conditional independence assumptions
and the assumption that $\mB$ is a deterministic function
of $\mIm$ yields
\begin{align}
  \sum_{\tilde \my}
  P_{\my \mid \hat \my, \mB_\mIm}(\tilde \my)
  \sum_{\mx \in \mX}
  H(\delta_{\tilde \my(\mx)} \,, \mQ_{\mtheta, \mIm}(\mx)),
\end{align}
where $\mB_\mIm$ denotes the boundaries as a function of $\mIm$.
Assuming that $y(x)$ and $y(x')$ are conditionally independent 
given $\mhaty, \mB$, $\forall \mx \neq \mx' \in \mX$ allows us 
to represent $P_{\my \mid \mhaty, \mB_{\mIm}}$ as a product 
of factors:
\begin{align}
  \sum_{\tilde \my}
  \left(
  \prod_{\mx' \in \mX} P_{\my(\mx') \mid \hat \my, \mB_\mIm}(\tilde \my(\mx'))
  \right)
  \sum_{\mx \in \mX}
  H(\delta_{\tilde \my(\mx)} \,, \mQ_{\mtheta, \mIm}(\mx)).
\end{align}
Factorizing out sums equal to one results in
\begin{align}
  \sum_{\mx \in \mX} \sum_{\tilde y(x)}
  P_{\my(\mx) \mid \mhaty, \mB_{\mIm}}(\tilde \my(\mx))
  H(\delta_{\tilde \my(\mx)} \,, \mQ_{\mtheta, \mIm}(\mx)).
\end{align}
Expanding the definition of cross-entropy yields the desired
expression:
\begin{align}
  \sum_{\mx \in \mX} 
  H(P_{\my(\mx) \mid \mhaty, \mB_\mIm}, \, \mQ_{\mtheta, \mIm}(\mx))
\end{align}

\subsection{Fr\'{e}chet derivative of matrix inverse}

This result is used in the derivation of the derivative of 
the random-walk partition function.  An intuitive derivation
is provided here.

Suppose $Ax=b$. We wish to find an expansion linear in $\epsilon V$
for $\tilde x := (A + \epsilon V)^{-1}b$, assuming the inverse exists.
\begin{align}
  (A + \epsilon V) \tilde x & = b\\
  A \tilde x & =  b - \epsilon V \tilde x\\
  \tilde x & = A^{-1}b - \epsilon A^{-1} V \tilde x\\
& = A^{-1}b - \epsilon A^{-1} V (A^{-1} b - O(\epsilon))\\
& = A^{-1}b - \epsilon A^{-1} V A^{-1} b + O(\epsilon^2)
\end{align}
where the second-to-last line follows from recursive expansion of the
same expression.
